\documentclass{article}
\usepackage[square,numbers]{natbib}
\usepackage[final]{nips_2017}
\usepackage{nips_2017}
\usepackage[english]{babel}

\usepackage[nottoc]{tocbibind}

\usepackage[utf8]{inputenc} 
\usepackage[T1]{fontenc}    
\usepackage{hyperref}       
\usepackage{url}            
\usepackage{booktabs}       
\usepackage{amsfonts}       
\usepackage{nicefrac}       
\usepackage{microtype}      
\usepackage{graphicx}
\graphicspath{{img/}}
\usepackage{adjustbox}

\title{Deep Learning for Metagenomic Data: using 2D Embeddings and Convolutional Neural Networks}

\author{
Nguyen Thanh Hai\\
UPMC- Paris 6 University\\ 
\texttt{nthai@cit.ctu.edu.vn} \\
\And
   Yann Chevaleyre\\
    Dauphine, PSL Research University, LAMSADE\\
   \texttt{yann.chevaleyre@dauphine.fr} \\
     \And
   Edi Prifti \\  
   Integromics, ICAN, Paris \\
   \texttt{e.prifti@ican-institute.org} \\
   \And
   Nataliya Sokolovska\\
   UPMC- Paris 6 University\\ 
   \texttt{nataliya.sokolovska@upmc.fr} \\
   \And
   Jean-Daniel Zucker \\  
   IRD, UMMISCO, France\\
   \texttt{jean-daniel.zucker@ird.fr} \\
}

\begin{document}

\maketitle

\begin{abstract}
Deep learning (DL) techniques have had unprecedented success when applied to images, waveforms, and texts to cite a few. In general, when the sample size ($N$) is much greater than the number of features ($d$), DL  outperforms previous machine learning (ML) techniques, often through the use of convolution neural networks (CNNs). However, in many bioinformatics ML tasks, we encounter the opposite situation where $d$ is greater than $N$. In these situations, applying DL techniques (such as feed-forward networks) would lead to severe overfitting. Thus, sparse ML techniques (such as LASSO e.g.) usually yield the best results on these tasks. In this paper, we show how to apply CNNs on data which do not have originally an image structure (in particular on metagenomic data). Our first contribution is to show how to map metagenomic data in a meaningful way to 1D or 2D images. Based on this representation, we then apply a CNN, with the aim of predicting various diseases. The proposed approach is applied on six different datasets including in total over 1000 samples from various diseases. This approach could be a promising one for prediction tasks in the bioinformatics field.\end{abstract}

\section{Introduction}

High throughput data acquisition in the biomedical field has revolutionized research and applications in medicine and biotechnology. Also known as omics data, they reflect different aspects of system's biology (genomics, transcriptomics, metabolomics and proteomics but also whole biological ecosystems acquired with the use of metagenomics). These datasets are increasingly available and numerous models use this information to make medical decisions \cite{Genomic_applications} (personalizing health care \cite{meta_personnal_medicine}, diagnosis and prognosis \cite{ML_large_metagenomic_tool_insight2016}, pharmacogenomics \cite{pharmacogenetics_2005}, etc.). However, exploring omics data has met many challenges (large number of features - genes/species $d$, and few observations $N$). Up to now, the most successful techniques applied to omics datasets have been mainly Random Forests (RF), and sparse regression. 
In this paper, we applied DL directly on six metagenomic datasets which reflect bacterial species abundance and presence in the gut of diseased patients and healthy controls. Since this technology performs particularly well in image classification, we focus here in the use of CNNs. In this context, it is important to find representations of the data that would be biologically pertinent to apply CNN techniques in order to learn new representations that would be used for classification purposes. Our objectives are to propose efficient representations which are compact in images, and to prove DL techniques as efficient tools for prediction tasks in the context of metagenomics.


There are numerous studies where authors have applied ML for analyzing large metagenomic datasets. Pasolli1 et al. \cite{ML_large_metagenomic_tool_insight2016} proposed a unified methodology to compare different state of the art methods (SVM, RF, etc) in various metagenomic datasets, which were processed with the same bioinformatics pipeline for comparative purposes. Authors in \cite{Deep_Learning_Health_Informatics_2017} introduced an overview of technologies in DL, which emerged and grew rapidly in recent years and have been applied to health informatics. Additionally, Zhao Y et al. in \cite{Exploration_machine_learning_sclerosis_disease_2017} leveraged ML methods to investigate prediction tasks for multiple sclerosis disease outcomes. Some studies applied ML techniques to predict dementia \cite{dementia_prognosis_2017} and cancer \cite{cancer_prognosis_and_prediction_2015}. Yoshua B. stated \cite{Bengio:2013:RLR:2498740.2498889_Representation_review} that the performance of prediction tasks depends on selecting representations, numerous studies have attempted to propose efficient representations. Montavon et al \cite{Montavon:2012:LIR:2999134.2999184} presented invariant representations of molecules to perform prediction tasks on atomization energy. The authors in \cite{fioravanti_phylogenetic_2017} introduced Ph-CNN using CNNs applying to metagenomic data embed the phylogentic tree.

Image classification based on ML has achieved impressive results and has been performing better than human experts in numerous cases. 
Yann Lecun et al. \cite{CNN_first_1989} proposed LeNet-5 as one of the first standard architecture for CNNs, whereas AlexNet as one of the most remarkable CNN architectures, showed significant improvements upon previous architectures. 
Since AlexNet, numerous architectures were introduced to improve the performance. Some of the most famous architectures among them are ZFNet \cite{ZFNet_2013}, GoogLeNet \cite{GoogLeNet_2014}, VGGNet \cite{VGGNet_2014}, ResNet \cite{ResNet_2015}, and WELDON \cite{durand_weldon:_2016}. A broad survey of CNNs was presented in \cite{recent_advance_cnn_2015} as proven that CNNs have gained a central position in image classification. 

\section{Methodology}\label{Methodology}

We evaluated our method with each of the different representations in the six datasets, which include bacterial species related to various disease, namely: liver cirrhosis (CIR), colorectal cancer (COL), obesity (OBE), inflammatory bowel diseases (IBD) and Type 2 Diabetes (T2D) \cite{ML_large_metagenomic_tool_insight2016,cirrhosis_2014,colorectal_2014,obesity_2013,ibd_2010,t2d_2012,wt2d_2013}, with CIR (n=232 samples with 118 patients), COL (n=48 patients and n=73 healthy individuals), OBE (n=89 non-obese and n=164 obese individuals), IBD (n=110 samples of which 25 were affected by the disease) and T2D (n=344 individuals of which n=170 are T2D patients). In addition, WT2 (n=96 European women with n=53 T2D patients and n=43 healthy individuals). The abundance datasets (ABD) are transformed to obtain another representation based on feature presence (PRE) when the abundance is greater than zero. These data were obtained using the default parameters of MetaPhlAn2 \cite{metaphan2} as detailed in Pasolli1 et al. \cite{ML_large_metagenomic_tool_insight2016}. For each sample, species abundance is represented as a real number and the total abundance of all species sums to 1.
\begin{figure}[ht]
\centering
\includegraphics[width=1\linewidth]{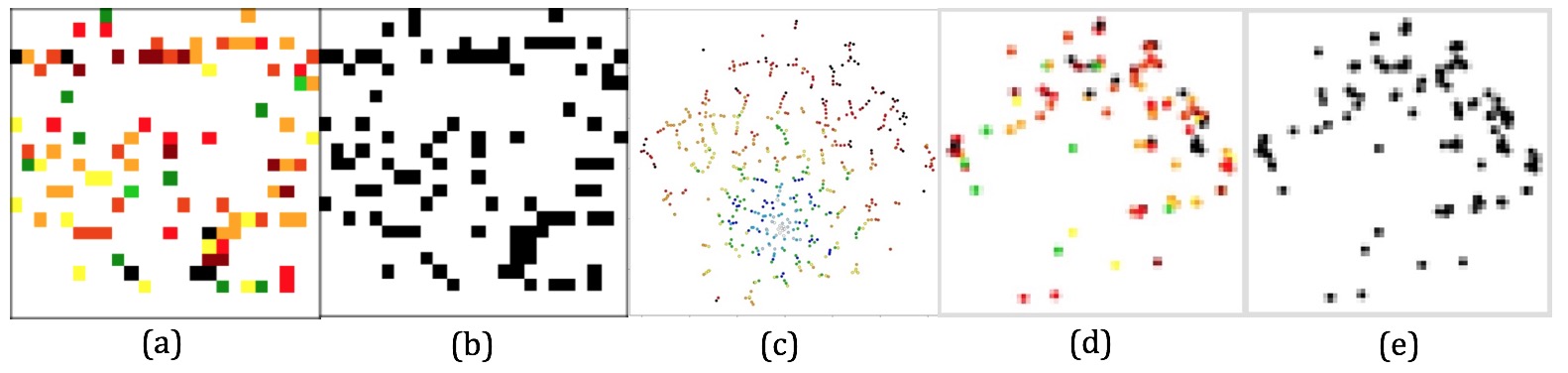}
\caption{Visualization of image-based representations. (a) square filling up from left to right/top to bottom with species abundances and a phylogenetic ordering; (b) same as a) but with presence instead of abundance; (c) a global t-SNE map was generated from a training dataset (Cirrhosis dataset), color represents mean abundance; (d) t-SNE image of a particular sample based on the same map, color represents abundance levels (e) same as d) but using presence instead of abundance.}
\label{fig:representations_color_strip}
\end{figure}
Our approach consists of the following steps: First, a set of colors is chosen and applied to different bins based on the abundance distribution. The binning can be performed on a linear or logarithmic scale. Here we present results from the latter. Then, the features are visualized into images by one of two different ways (\textbf{phylogenetic-sorting} (PLG) or visualized based on t-Distributed Stochastic Neighbor Embedding (\textbf{t-SNE}) \cite{tsne_2008}). t-SNE technique is so useful to find faithful representations for high-dimensional points visualized in a compact space, typically the 2D plane. For phylogenetic-sorting, the features which are bacterial species are arranged based on their taxonomical annotation ordered increasingly by the concatenated strings of their taxonomy (i.e. phylum, class, order, family, genus and species). This ordering of the variables integrates within the data an external biological knowledge, which reflects the evolutionary relationship between these bacterial species. Each visualization method will be used to either represent abundance or presence data. A fifth representation, which serves as control is the 1D of the raw data (with the species also sorted phylogenetically).
For the t-SNE approach, we use only training sets to generate global t-SNE maps, images of training and test set are created from these global maps. The representations are evaluated in a framework with 100 CNN architectures and a network with a fully connected layer without convolutions. We used accuracy (ACC) to measure model performances. Results represent average accuracy values through a 10-fold cross-validation. All the above mentioned architectures are implemented in Torch7 \cite{torch7}. 



\subsection{Approaches to generate images: Fill up and t-SNE}
\textbf{“Fill up”} images are created by arranging abundance/presence values into a matrix in a left-to-right order by row. The image is square and empty bottom-left of the image are set to zero (white). As an example for a dataset containing 542 features (i.e. bacterial species) in the cirrhosis dataset, we need a matrix of 24x24 to fill up 542 values of species into this square. The first row of pixel is arranged from the 1st species to the 24th species, the second row includes from the 25th to the 48th and so on till the end. We use distinct colors in a logarithmic binning scale to illustrate abundance values of species and black and white for presence/absence, where white represents absent values.

\textbf{T-SNE maps}: are built based on training sets from raw data with perplexity = 10 after 500 epochs. These maps are used to generate images for training and testing sets. 
Each species is considered as a point in the map and only species that are present are showed either in abundance or presence using the same color schemes as above. Figure \ref{fig:representations_color_strip} illustrates images generated by "fill-up" and t-SNE.

\subsection{Architecture configurations of CNNs}
Our proposed architectures are mainly inspired by the philosophy of VGG nets \cite{VGGNet_2014}, including 3x3 filters with stride 1, and max pooling of 2x2 with stride 2, using ReLU after each convolution. In the experiments, image dimensions range from 32x32 for fill up and 64x64 for t-SNE. These images are passed through a stack of convolutional layers (the depth ranging from one to five convolutional layers, and the width of each convolutional layer increasing from one, up to twenty filters), followed by one max pooling, and one Fully-Connected layer (\textbf{FC}). 
\begin{figure}[ht]
\centering
\includegraphics[width=1\linewidth]{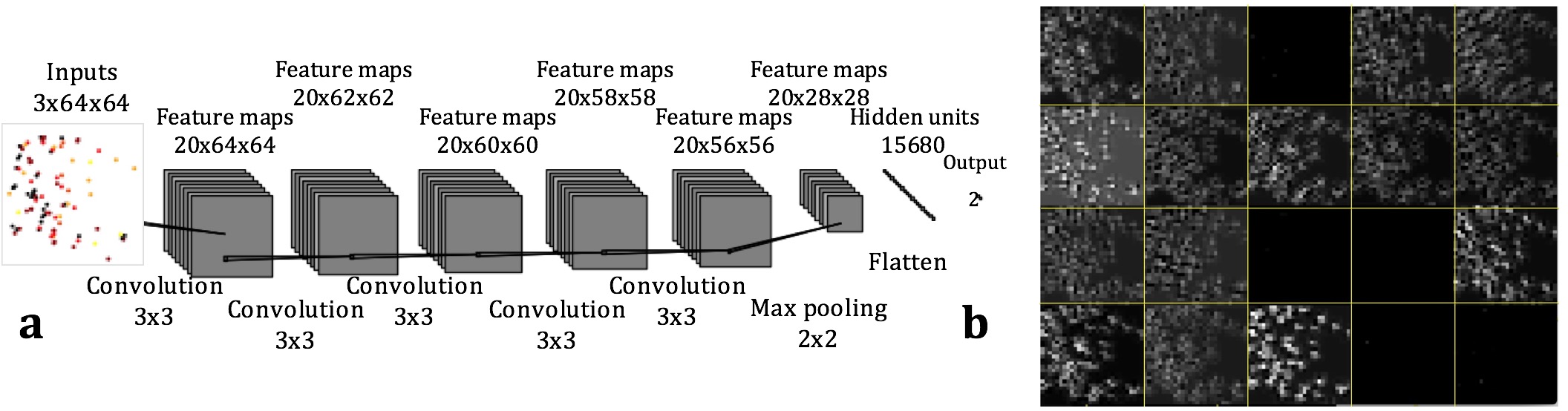}
\caption{An illustration of the CNN architecture for representations based on synthetic images. \textbf{(a)} The architecture receives a three-channel color image as input passing a stack of five convolutional layers (with 20 kernels of 3x3 (stride 1) for each), followed by a max pooling 2x2 (stride 2). The last layer is fully connected, taking feature maps from a stack of convolutional and max pooling layers as input in a vector form, and computes the scores for the output by LogSoftMax function. \textbf{(b)} Visualization of 20 feature maps generated after a stack of convolutional layers and max pooling of a fully trained model (the architecture (a)) with the input image at (a).}.
\label{fig:example_cnns}
\end{figure}
For the first convolutional layer, wide convolution \cite{Kalchbrenner14aconvolutional} is implemented to preserve the dimension of input. Besides 2D convolutions for images, we also apply a 1D convolution to raw data with kernel 3x1 (step 1), max pooling with size 2, stride 2. There are two approaches for binary classification including using two output neurons (two-node technique) or one output neuron (one-node technique). For the two-node approach, the final layer is the LogSoftMax layer with 2 outputs, one indicating patient was affected by the disease, the other shows the patient was unaffected (see an example in Figure \ref{fig:example_cnns}). For the one-node approach, we used a sigmoid activation function at the final layer and binary cross-entropy for Sigmoid. In our experiments, the two-node approach with LogSoftMax gave better results, so the architectures in Table \ref{tab-res} used this method. The networks are trained using stochastic gradient descent for the two-node approach or Adam \cite{adam_alg_2014} for the one-node approach with a mini-batch size of 16, momentum of 0.1, and weight decay of $10^{-5}$. Constant learning rate is set at 0.0005. 

\section{Results}
\begin{table}[t]
\caption{Prediction performance (in ACC) of representations after 200 epochs. The results that are better the state of the art ML approaches are formatted in \textbf{bold}. PLG, t-SNE are ways to arrange species. ABD, PRE are two types of features (ABD: species abundance, and PRE: marker presence). (1) are results of the neural network with a fully connected layer without convolutions (FC), (2) indicates performance of CNN architectures (Con1D: 1D convolutions, Con2D: 2D convolutions). MetAML \cite{ML_large_metagenomic_tool_insight2016} is a computational tool to perform prediction tasks on metagenomic data using machine learning classifiers such as support vector machines (SVM), RF, Lasso, and Elastic Net. We selected the best classifiers of MetAML including RF, and SVM for comparison. }
  \label{tab-res}
  \centering
  \begin{tabular}{lllllllllll}
    \toprule
     Datasets &  PLG & PLG & PLG  & t-SNE  & t-SNE &\multicolumn{2}{c}{MetAML} \\
       & Raw & ABD  & PRE &  ABD & PRE \\
     & Con1D & Con2D & Con2D  & Con2D & Con2D & RF & SVM  \\
    \midrule
    CIR(1) & 0.852  & \textbf{0.878} & 0.865  & 0.743 & 0.817         & 0.877 & 0.834 \\
    CIR(2)& 0.870  &   \textbf{0.891} & 0.874 & 0.865 &  0.852 \\
    COL(1)& 0.675 &  0.767 & 0.775 & 0.650 & 0.608       & 0.805& 0.743 \\
    COL(2)& 0.717  & 0.742 & 0.725 & 0.783 & 0.708  \\
    IBD(1)& \textbf{0.818}  &  \textbf{0.855} & \textbf{0.827}  & 0.682 & 0.755 & 0.809      & 0.809 \\
    IBD(2)&   0.763    & \textbf{0.836} & 0.764 & 0.809 & \textbf{0.818}  \\
    OBE(1)& 0.588  &  0.636 & \textbf{0.656}  & 0.628 & 0.596            & 0.644 & 0.636\\
    OBE(2)&  0.616     &  \textbf{0.660} & 0.592 & 0.628 & 0.628  \\
    T2D(1)& 0.635  &  0.662 & 0.638  & 0.565 & 0.588        & 0.664  & 0.613 \\
    T2D(2)&   0.639    & 0.626 & 0.656 & 0.635 & 0.647  \\
    WT2(1)& 0.656  &  0.678 & \textbf{0.733} &0.611 &  0.622 & 0.703 & 0.596\\
    WT2(2)&   0.556  &  0.589 & 0.633 & \textbf{0.711} & \textbf{0.711} \\
    \bottomrule
  \end{tabular}
\end{table}

Table \ref{tab-res} illustrates the results of the CNN architecture with 5 convolutional (consisting of 2D convolutions) layers and 20 filters per layer to images-based representations and the convolutional architecture including 2 convolutional layers (1D convolutions) and 20 filters per layer for the raw data. IBD dataset illustrated the greater improvement with most representations performing better than state of the art results. Images-based representations are more efficient than raw data, while 2D convolutions also surpasses 1D convolutions. For the t-SNE representation, the CNN architecture outperforms FC.
\section{Conclusion}
We proposed the MET2IMG approach to predict patients' diseases using metagenomic data. Our method is very promising in the context of prediction using metagenomic data. We used two main approaches to build "synthetic images" including the "fill-up" and the t-SNE embedding. For the "fill-up" approach, we are able to use smaller and simple images. For the t-SNE representations,  features are embedded in a two-dimensional space using a classic embedding approach in ML. The dimension of t-SNE images is required to be higher to get a higher performance because some data points might be overlapped with a small scaled dimension, while each feature in the "fill-up" approach is visible. Hence, learning with t-SNE images may be more complicated as well as require more time and more memory to process compared to "fill-up". We only evaluated small images with dimensions of 32x32 and 64x64 due to limitations of computational resources, so the deeper architectures of CNNs should be investigated to evaluate on larger-scaled images. Besides, further research should compare the performance of image-based representations against tree-based representations. 
\bibliography{sample}
\end{document}